\documentclass[letterpaper, 10 pt, conference]{ieeeconf}  

\IEEEoverridecommandlockouts                              

\usepackage{graphics} 
\usepackage{amsmath} 
\usepackage{epsfig} 
\usepackage{mathptmx} 
\usepackage{times} 
\usepackage{amsmath} 
\usepackage{amssymb}  
\usepackage[dvipsnames]{xcolor}
\usepackage{booktabs}
\usepackage{multirow}
\usepackage{makecell}
\usepackage{pifont} 
\usepackage{colortbl}
\usepackage{amssymb}

\definecolor{darkgreen}{RGB}{59,125,35}
\definecolor{light_purple}{RGB}{216,110,204}
\definecolor{dark_blue}{RGB}{8, 79, 106}
\definecolor{dark_purple}{RGB}{102,0,102}
\definecolor{Grey}{rgb}{0.5, 0.5, 0.5}

\title{\LARGE \bf
Fast-SmartWay: Panoramic-Free End-to-End Zero-Shot Vision-and-Language Navigation}

\author{Xiangyu Shi$^1$, Zerui Li$^1$, Yanyuan Qiao$^{2\mathsection}$, Qi Wu$^{1\dagger}$
\thanks{$^1$Xiangyu Shi, Zerui Li, and Qi Wu are with the Australian Institute for Machine Learning, the University of Adelaide. Xiangyu Shi ({\tt\footnotesize xiangyu.shi@adelaide.edu.au})} 
\thanks{$^2$Yanyuan Qiao is with the CREATE Lab, Swiss Federal Institute of Technology Lausanne (EPFL)}
\thanks{$\mathsection$ Project Lead: Yanyuan Qiao ({\tt\footnotesize yanyuan.qiao@epfl.ch})} 
\thanks{$\dagger$ Corresponding author: Qi Wu ({\tt\footnotesize qi.wu01@adelaide.edu.au})}
}

\begin{document}

\maketitle
\thispagestyle{empty}
\pagestyle{empty}

\begin{abstract}
Recent advances in Vision-and-Language Navigation in Continuous Environments (VLN-CE) have leveraged multimodal large language models (MLLMs) to achieve zero-shot navigation. However, existing methods often rely on panoramic observations and two-stage pipelines involving waypoint predictors, which introduce significant latency and limit real-world applicability. In this work, we propose Fast-SmartWay, an end-to-end zero-shot VLN-CE framework that eliminates the need for panoramic views and waypoint predictors. Our approach uses only three frontal RGB-D images combined with natural language instructions, enabling MLLMs to directly predict actions. To enhance decision robustness, we introduce an Uncertainty-Aware Reasoning module that integrates (i) a Disambiguation Module for avoiding local optima, and (ii) a Future-Past Bidirectional Reasoning mechanism for globally coherent planning. Experiments on both simulated and real-robot environments demonstrate that our method significantly reduces per-step latency while achieving competitive or superior performance compared to panoramic-view baselines. These results demonstrate the practicality and effectiveness of Fast-SmartWay for real-world zero-shot embodied navigation.
\end{abstract}

\section{INTRODUCTION}

The Vision-and-Language Navigation (VLN) task~\cite{anderson2018vision} aims to enable embodied agents to navigate in environments based on natural language instructions~\cite{anderson2018vision,wang2023scaling}. Early VLN approaches~\cite{hong2020recurrent,chen2021hamt,qiao2022hop} typically relied on pre-defined navigation graphs, which constrained the agent’s decision-making within a static environment representation and limited real-world applicability. To address these limitations and improve the deployability of VLN in realistic scenarios, a Vision-and-Language Navigation in Continuous Environment (VLN-CE) benchmark~\cite{krantz2020beyond,hong2022bridging} was proposed. Unlike previous tasks, VLN-CE removes the dependence on pre-constructed maps by introducing a waypoint predictor~\cite{hong2022bridging} to dynamically identify traversable regions. This mechanism empowers the agent to operate in previously unseen environments without prior knowledge of the layout, making the task more practical and generalizable.

Traditional VLN-CE systems~\cite{an2022bevbert, an2024etpnav, li2025ground,krantz2022sim} are usually trained on large-scale datasets, learning to map language instructions and visual observations into a sequence of navigation actions. While these trained agents may perform well in seen or similar environments, they often struggle to generalize to unfamiliar scenarios due to overfitting to specific visual or structural patterns. This challenge has led to an increasing interest in integrating large language models (LLMs)~\cite{touvron2023llama} and multimodal large models (MLLMs)~\cite{openai2023gpt4} into both VLN and VLN-CE area. For instance, NavGPT~\cite{zhou2024navgpt} is one of the first attempts to use GPT-4~\cite{openai2023gpt4} in zero-shot VLN by converting visual inputs into textual descriptions and feeding them into GPT-4 for reasoning and direction prediction. Open-Nav~\cite{opennav} first utilises open-source LLMs in a zero-shot VLN-CE setting by employing a waypoint predictor~\cite{hong2022bridging} to generate candidate navigable locations, converting each candidate's associated visual input into a textual description, and then feeding these descriptions into the LLMs for instruction interpretation and navigation decision making. In contrast, SmartWay~\cite{smartway} leverages GPT-4o’s native multimodal capabilities, directly feeding images and text into the model for joint visual-linguistic reasoning. 
\begin{figure}[t]
    \centering
    \includegraphics[width=\linewidth]{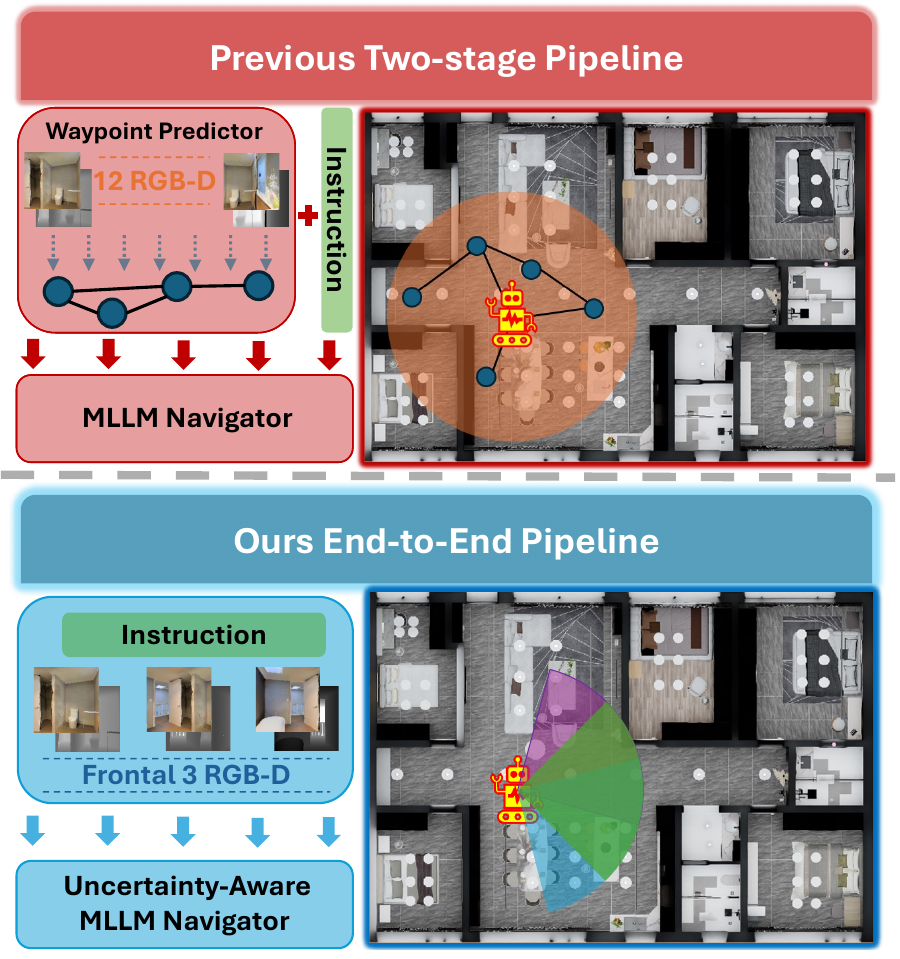}
    \vspace{-20pt}
    \caption{Comparison between previous VLN-CE pipelines and our proposed end-to-end framework. \textbf{Top:} Previous two-stage methods first feed panoramic RGB-D observations (12 views) into a waypoint predictor to generate candidate waypoints. In the second stage, the candidate waypoints, together with the instruction, are provided to the navigator to select the final navigation action. \textbf{Bottom:} Our framework integrates the instruction with three frontal-view images in an Uncertainty-Aware Navigator, directly predicting navigation actions in a single end-to-end step.}
    \label{fig:reasoning-process}
\end{figure}

Although recent zero-shot VLN-CE methods using LLMs and MLLMs have shown promising results, they still face key challenges in real-world deployment. Most of these methods rely on panoramic observations, which require either rotating the robot 360 degrees to capture 12 views or using specialized panoramic cameras. This process is slow, introduces latency in dynamic environments, and is often incompatible with compact robots that only have a front-facing camera. 
In addition, many existing methods~\cite{opennav,smartway} rely on waypoint predictors that select navigable points based only on RGB-D inputs, without considering the semantic alignment between the current environment and the instruction. As a result, the selected candidates may not align with the intended goal, leading to suboptimal decisions. This reliance on discrete candidates also prevents the agent from generating smooth and continuous trajectories, further limiting their practicality in real-world navigation tasks.

To address these limitations, we propose an End-to-End VLN-CE framework named \textbf{Fast-SmartWay}, shown in Fig.~\ref{fig:reasoning-process}, that significantly reduces the processing time for each action step. Unlike prior methods that rely on panoramic observations, typically requiring 12 images captured by rotating 30 degrees each in traditional VLN-CE settings, we leverage only the front three-view images as visual input. These images, together with the language instruction, are directly fed into a multimodal large model (MLLM), which predicts the turning angle and forward distance in a single forward pass, enabling fully end-to-end navigation.

In addition, we propose a \textbf{Spatial-Semantic Textual Description Generation} module that replaces traditional waypoint predictors by transforming spatial layouts and semantic cues into textual prompts. These prompts, combined with the language instruction, are directly fed into the MLLM to predict navigation actions, eliminating the need for explicit candidate selection. This design tightly integrates perception and decision-making, enhancing both efficiency and generalisation.

Furthermore, we introduce an \textbf{Uncertainty-Aware Reasoning} module to enhance the robot's robustness and decision quality in challenging environments. It consists of two components: (1) a Disambiguation module, which dynamically evaluates the diversity and uncertainty of path choices to help the robot avoid dead-ends and local optima, and (2) inspired by NavBench's~\cite{qiao2025navbench} Local Observation ability, we propose a Future-Past Bidirectional Reasoning (FPBR) mechanism, which leverages both historical analysis and future predictions to guide the MLLM in evaluating the rationality of its current decisions, improving the robot’s adaptability and success rate in dynamic environments without requiring additional training.

Our main contributions are as follows:
\begin{itemize}
  \item We present an \textbf{End-to-End VLN-CE Framework} that eliminates the need for panoramic observations and waypoint predictors by leveraging only {frontal views}, enabling more efficient end-to-end navigation.
  \item We propose an Uncertainty-Aware Reasoning module, which integrates a Disambiguation module for avoiding dead-ends and a FPBR mechanism for enhancing planning consistency and robustness.
  \item We conduct extensive experiments on standard VLN-CE benchmarks, demonstrating the effectiveness and practicality of our method in zero-shot settings.
\end{itemize}

\begin{figure*}
    \centering
    \includegraphics[width=1\linewidth]{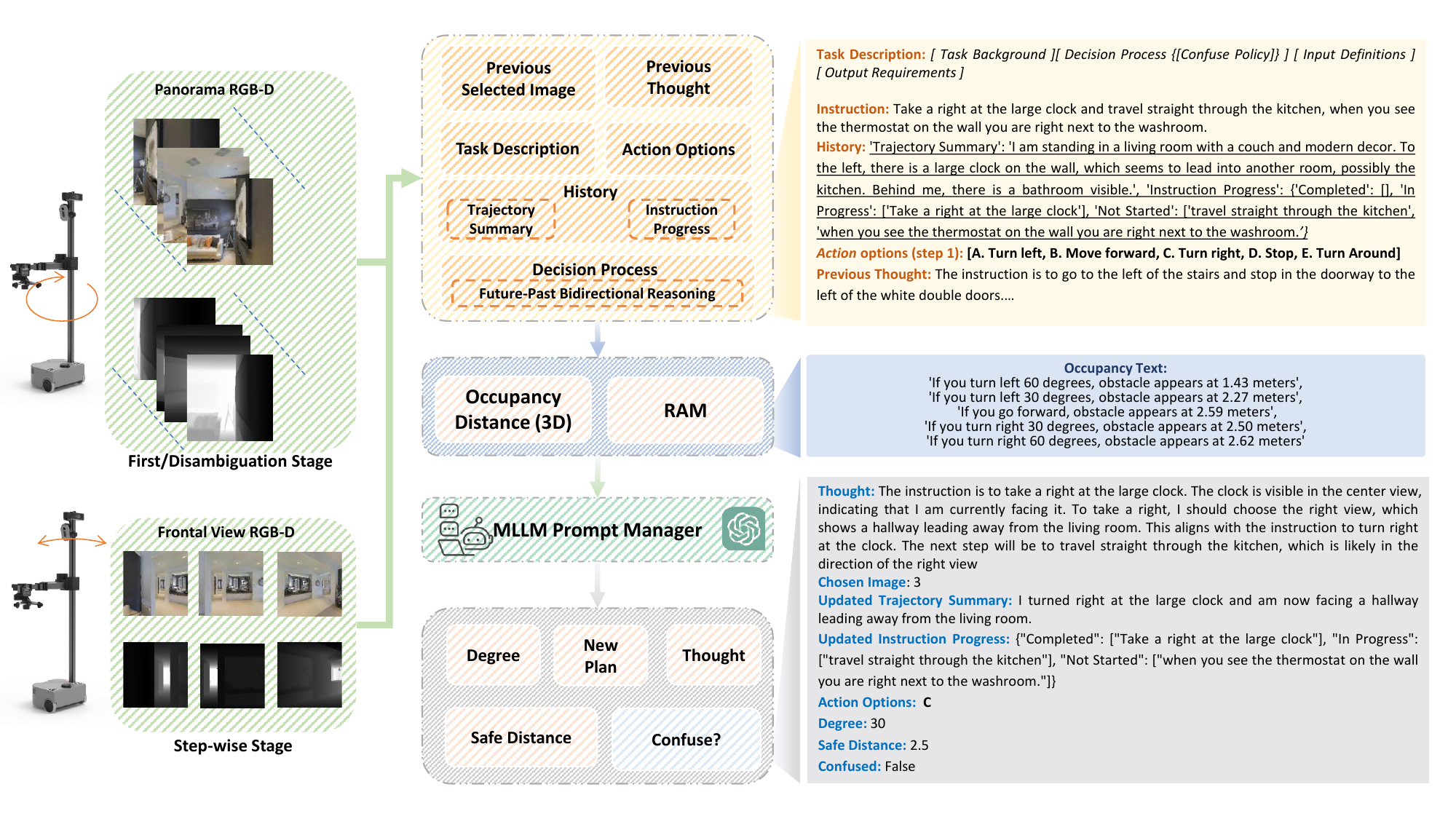}
    \vspace{-20pt}
    \caption{Overall workflow of our proposed zero-shot VLN-CE framework. The navigation process begins with panoramic RGB-D observations at the initial or disambiguation stage, and uses three frontal RGB-D views during the step-wise stage. The system first constructs structured prompts, then extracts spatial and semantic descriptions, and finally sends both together to the Multimodal Large Language Model (MLLM). Within the Decision Process, the MLLM performs step-wise reasoning and incorporates Future-Past Bidirectional Reasoning (FPBR) to ensure globally consistent planning. It also determines whether the robot is confused based on the instruction and context. If so, the Disambiguation Module is triggered to collect a panoramic observation and replan. The robot used in the figure is a Hello Robot~\cite{HelloRobot2025} equipped with an Intel RealSense camera mounted at a height of 125 cm.}
    \label{fig:overall_diagram}
\end{figure*}

\section{Related Work}
\label{sec:related work}
\subsection{Vision-and-Language Navigation}
Vision-and-Language Navigation (VLN) tasks require agents to follow instructions to reach target locations in unseen environments~\cite{anderson2018vision,zhang2024vision}. Early benchmarks are built on discrete environments with predefined navigation graphs~\cite{qi2020reverie,anderson2020rxr,thomason2020cvdn}. To improve realism, Krantz et al.~\cite{krantz2020beyond} extended VLN to continuous environments (VLN-CE), enabling agents to navigate in more realistic, map-free settings. To support this, subsequent works~\cite{hong2022bridging} introduced waypoint predictors to estimate traversable candidates based on the agent’s current observation. These predictors have since become a core component in many VLN-CE systems~\cite{an2022bevbert,an2024etpnav}. However, training such systems remains data-intensive and environment-specific, leading to growing interest in zero-shot VLN approaches that leverage the generalization capabilities of large foundation models.

\subsection{Zero-Shot VLN}
To reduce reliance on task-specific data and improve generalization to unseen environments, recent studies explore zero-shot VLN approaches by leveraging the powerful reasoning capabilities of large language models (LLMs) and multimodal large models (MLLMs). These methods aim to bypass the need for environment-specific training by prompting pretrained models with visual and linguistic inputs. NavGPT~\cite{zhou2024navgpt} is an early attempt that converts visual observations into textual descriptions and feeds them, along with the instruction, into GPT-4 for navigation reasoning. DiscussNav~\cite{long2023discuss} further adopts a multi-expert framework, assigning the VLN task to GPT4 for execution. Open-Nav~\cite{opennav} extends this idea to the VLN-CE setting by combining a learned waypoint predictor with open-source LLMs, which evaluate each candidate location based on its textual description. More recently, SmartWay~\cite{smartway} utilizes GPT-4o's multimodal ability, enabling visual-linguistic reasoning with a backtrack mechanism. Nevertheless, these methods often rely on panoramic inputs and incur significant processing latency, motivating the exploration of frontal-view VLN settings.

\subsection{Ego-Centric  VLN}
Most existing VLN and VLN-CE methods rely on panoramic observations obtained through 360° rotations or specialized sensors. However, such configurations are often impractical for real-world robots, which typically operate with only a forward-facing camera. Recent efforts, such as Navid~\cite{zhang2024navid}, have explored ego-centric settings by leveraging sequences of past front-view observations combined with the latest frame for decision-making. Similarly, Wang et al.~\cite{wang2024sim} propose building a semantic traversability map to identify navigable regions and guide agents along feasible paths. Nevertheless, ego-centric perception can cause agents to remain on incorrect trajectories without realizing deviations from the instruction, sometimes following a wrong route until the end. To address this challenge, our work employs three forward-facing views and introduces an uncertainty-aware reasoning module, enhancing robustness perception without additional training.

\section{Problem Formulation}
\label{sec:problem_formulation}

In this work, we tackle the problem of \textit{Vision-and-Language Navigation in Continuous Environments} (VLN-CE), where an agent needs to navigate through a 3D environment \(\mathbf{E}\) following the natural language instruction \(L = \{l_1, l_2, \dots, l_n\}\).

\paragraph{Classical VLN-CE Setting.} In prior VLN-CE methods, the agent at each time step \(t\) is assumed to have access to a full panoramic observation. This panorama consists of 12 RGB-D images captured at evenly spaced headings (e.g., \(0^\circ, 30^\circ, \dots, 330^\circ\)), yielding:
$I_{t} = \{ (I^{rgb}_i, I^{depth}_i) \mid i = 1, \dots, 12 \}$,
where each \(I^{rgb}_i \in \mathbb{R}^{H \times W \times 3}\) and \(I^{depth}_i \in \mathbb{R}^{H \times W}\). However, such 360-degree perception sensing is often challenging in real-world scenarios due to hardware limitations and time constraints for multi-view scanning.

\paragraph{Our Setting} To better reflect realistic constraints, we propose a more practical setting where the agent only performs one panoramic scan at the initial step \(t=0\), yielding the initial input:
$
I_{0} = \{ (I^{rgb}_i, I^{depth}_i) \mid i = 1, \dots, 12 \}.
$
This scan provides a coarse environment understanding only before navigation begins. For all subsequent steps \(t > 0\), the agent receives only frontal observation consisting of 3 RGB-D views captured at heading angles \((330^\circ, 0^\circ, 30^\circ)\), corresponding to the left, front, and right directions:
$
I_{t} = \{ (I^{rgb}_i, I^{depth}_i) \mid i = 1, 2, 3 \}.
$

At each step, given the current partial observation \(I_t\) and instruction \(L\), the agent selects a low-level action (i.e., movement direction and distance) to progressively follow the instruction and reach the goal location \(\mathbf{x}_{\text{goal}}\).

\section{Method}

As illustrated in Fig.~\ref{fig:overall_diagram}, our method begins with an end-to-end, zero-shot VLN-CE pipeline. Unlike prior zero-shot approaches~\cite{opennav,smartway} that rely on waypoint predictors~\cite{hong2022bridging} to generate candidate turning angles and movement distances from panoramic views, our method utilizes a Multimodal Large Language Model (MLLM) to directly infer the robot’s next action, including both turning angle and forward distance based on the given instruction and the frontal view.

While this end-to-end formulation simplifies the navigation pipeline by eliminating intermediate heuristics, it may still encounter ambiguous or conflicting situations during navigation, especially in complex indoor environments. To enhance the robustness of our MLLM-based inference, we introduce an \textit{Uncertainty-Aware Reasoning} mechanism. This mechanism enables the robot to (i) identify and recover from uncertain cases via a Disambiguation Module, and (ii) incorporate both past trajectories and anticipated future subgoals to enable globally consistent path planning.
\subsection{End-to-End Pipeline}
In prior zero-shot VLN-CE methods~\cite{opennav,smartway}, the waypoint predictor is adopted as an intermediate module to process a panoramic RGB-D input, which is represented by 12 distinct images captured at 30-degree intervals. The module subsequently generates a discrete set of waypoint candidates, each of which is presented as an image for the navigator to make a selection. However, this two-stage design introduces a structural disconnection between vision and language. The waypoint predictor is typically trained only by RGB-D, without an instruction understanding module. As a result, the generated candidates may be visually plausible but semantically irrelevant to the goal, leading to suboptimal choices. 

In contrast, under our frontal view setting, instead of retraining a waypoint predictor to operate on restricted-view observations, we completely remove the waypoint predictor and propose the End-to-End Navigator that asks MLLM to directly generate the next action.

\noindent\textbf{Spatial-Semantic Textual Description Generation}:
To capture spatial information from the robot’s depth observations, we construct a partial panoramic point cloud using three depth images
 $\{ I^{depth}_i \mid i \in \{L, F, R\} \}$, corresponding to views taken at $30^\circ$ to the left, forward, and $30^\circ$ to the right of the robot’s heading. We extract a centre crop from each image to focus on the most reliable region, thereby reducing peripheral distortion and improving depth quality. To focus on nearby obstacles on the ground, we retain only the bottom half of each depth map during projection. Each cropped depth image is then projected into a 3D space using the pinhole camera model.  The resulting local point clouds are finally transformed into the world frame using the robot’s pose and the relative yaw of each view.

Let the resolution of each cropped depth image be $(h, w)$, where $h$ denotes the image height and $w$ the width. 
After the projection, each depth map yields a set of 3D points arranged in a $(h/2, w)$ grid, corresponding to the bottom half of the image. 
We then compute the ground-plane (horizontal) distance of each 3D point to the robot:
\[
D^{(i)}_j = \sqrt{x^2 + z^2},
\]
where $(x, z)$ are the 2D coordinates of a 3D point in the robot’s local frame, 
$i \in \{L, F, R\}$ indexes the view direction, and $j \in \{1, \ldots, w\}$ indexes the image column. 

To robustly estimate the distance to the nearest obstacle in each direction, we compute the minimum ground-plane distance within each column:
\[
D^{(i)}_j = \min_{k \in \{1, \ldots, h/2\}} \sqrt{x_{k,j}^2 + z_{k,j}^2}.
\]
This results in three 1D vectors $\{ D^{(L)}, D^{(F)}, D^{(R)} \}$, each of length $w$, encoding the closest ground-level obstacles in the left, front, and right views, respectively.

To make the distances interpretable for the MLLM, we concatenate the $\{ D^{(L)}, D^{(R)} \}$, covering a $120^\circ$ horizontal field of view ($-60^\circ$ to $+60^\circ$). 
This range is discretized into $N=5$ directional bins for five directions: turning left $60^\circ$, turning left $30^\circ$, going forward, turning right $30^\circ$, and turning right $60^\circ$. 
For each bin $b \in \{1, \ldots, 5\}$, we compute the mean distance $\bar{d}_b$ of all 3D points within its angular range and generate a textual description in the form \textit{``If you [direction], ...''}, where the direction is determined by the bin’s orientation.

To generate natural language descriptions, we compare each $\bar{d}_b$ against two thresholds, $d_{\text{close}}$ and $d_{\text{mid}}$, and select the textual description based on the following rule:

\[
\resizebox{\linewidth}{!}{$
\ell_b = \begin{cases}
\text{`If you [direction], there is a very close obstacle''}, & \text{if } \bar{d}_b < d_{\text{close}} \\
\text{`If you [direction], obstacle appears at } \bar{d}_b \text{ meters''}, & \text{if } d_{\text{close}} \leq \bar{d}_b < d_{\text{mid}} \\
\text{`If you [direction], path is clear for moving forward in } \bar{d}_b \text{ meters''}, & \text{otherwise}
\end{cases}
$}
\]

We denote the resulting set of spatial descriptions as \(\mathcal{L}_{\text{spatial}}^{(5)} = \{\ell_1, \ell_2, \ell_3, \ell_4, \ell_5\}\), where each $\ell_i$ corresponds to one of the 5 discretized orientations \(-60^\circ, -30^\circ, 0^\circ, +30^\circ, +60^\circ\).

The spatial description generation is reused not only during regular navigation steps, but also during two special situations: (i) the first navigation step, where the robot performs a 360° rotation to understand its full surroundings, and (ii) the Disambiguation stage, which is triggered when the robot encounters uncertain or conflicting signals. In both cases, the robot performs a $360^\circ$ turn and captures 12 depth images (rotate $30^\circ$ each) with a horizontal resolution of $w=256$. 
In this case, we compute distances only from the central half of each image (columns $[64,192]$) to reduce redundant obstacles. 
For each view $i \in \{1, \ldots, 12\}$, we calculate the mean distance $\bar{d}_i$ of the selected segment and generate a natural language statement based on its global orientation, such as 
\textit{``go forward''}, \textit{``turn left $90^\circ$''}, \textit{``turn right $90^\circ$''}, or \textit{``turn around''}. Directions falling within the left half of the panoramic view are described as ``turn left,'' while those exceeding $180^\circ$ are normalised and expressed as ``turn right'' with the equivalent angle.
Descriptions again indicate whether the path contains a very close obstacle, a mid-range obstacle with distance annotation, or a clear corridor of length. 
This ensures that the robot’s full surrounding context is expressed in textual form at the beginning of navigation. 
We obtain \(\mathcal{L}_{\text{spatial}}^{(12)} = \{\ell_1, \ell_2, \ldots, \ell_{12}\}\), where each $\ell_i$ describes the spatial information at orientation $30^\circ \times (i-1)$.

In addition to spatial descriptions, we incorporate semantic information from RGB observations. 
Specifically, we utilise the RAM model~\cite{Zhang2023RecognizeAA} to identify a set of explicitly detected semantic objects from each RGB image. 
For the frontal view setting, this yields three object lists 
$\mathcal{O}^{(3)} = \{\mathcal{O}_L, \mathcal{O}_F, \mathcal{O}_R\}$, 
corresponding to the left, front, and right RGB images, $I^{rgb}_{1...3}$. 
During the panoramic setting with 12 views, we obtain 12 object lists 
$\mathcal{O}^{(12)} = \{\mathcal{O}_1, \mathcal{O}_2, \ldots, \mathcal{O}_{12}\}$, 
where each $\mathcal{O}_i$ denotes the detected objects from the $i$-th RGB image in the sequence $I^{rgb}_{1...12}$.

\noindent\textbf{Multimodal Prompt Construction}: 
At the beginning of a navigation episode, the robot’s orientation is unknown and no actions have yet been executed. 
The input at the initial step consists of: 
(1) the instruction text $\mathcal{I}$. 
(2) semantic descriptions $\mathcal{O}^{(12)}$ and spatial descriptions $\mathcal{L}_{\text{spatial}}^{(12)}$, which together provide a comprehensive understanding of the scene to the MLLM.
(3) the task description $\mathcal{T}$, which defines the high-level objective and operational constraints of the navigation task. 
(4) the 12 panoramic RGB images $I^{rgb}_{1...12}$.
Given this multimodal input, the model reasons about the scene and the instruction to determine the most promising initial direction and movement. It produces the following structured outputs:
\begin{itemize}
    \item \textbf{Thought:} A natural language explanation of how the instruction and scene observations were interpreted to choose the initial direction.
    \item \textbf{Selected Image (1--12):} The index of the RGB image that best represents the direction the robot should initially face and begin moving toward.
    \item \textbf{Safe Distance:} A forward distance (in meters) the robot can safely travel in the selected direction, constrained by nearby obstacles.
    \item \textbf{Trajectory Summary:} A natural language overview describing the surrounding environment based on the panoramic observation.
    \item \textbf{Instruction Progress:} Ask $\mathcal{I}$ split into subgoals with status: first subgoal \texttt{In Progress}, the rest \texttt{Not Started}, and none \texttt{Completed}.
\end{itemize}

\noindent\textbf{Step-wise Multimodal Reasoning:} 
After the initial orientation is determined, the robot continues navigation with frontal view. The inputs include: 
(1) three RGB images from different views, namely left ($-30^\circ$), center ($0^\circ$), and right ($+30^\circ$). 
(2) textual navigation instruction $\mathcal{I}$, semantic descriptions $\mathcal{O}^{(3)}$ and spatial descriptions $\mathcal{L}_{\text{spatial}}^{(5)}$, and a list of valid high-level action options. 
(3) contextual information from the previous step, including the Observed objects $\mathcal{O}$,  
\texttt{Previous Selected Image} (\texttt{Selected Image} from last step), \texttt{Instruction Progress}, \texttt{Trajectory Summary}, and the \texttt{Previous Thought} that supported the last action.
\begin{figure*}
    \centering
    \includegraphics[width=\linewidth]{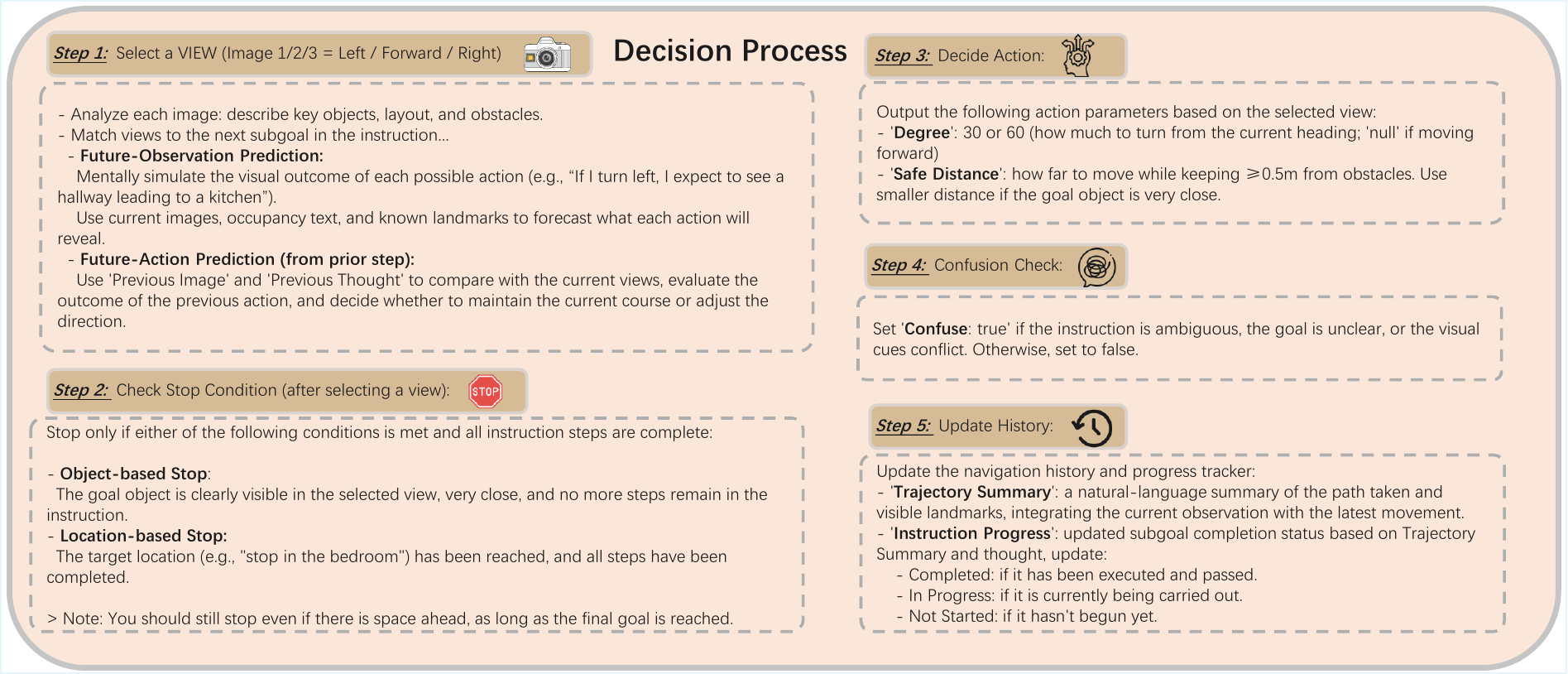}    
    \caption{Overview of the structured Decision Process in the MLLM-based Step-wise navigation pipeline. Upon receiving inputs, the model evaluates candidate views, predicts future observations, analyses the previous action, determines stop conditions, and estimates safe moving distance with selected direction.}
    \label{fig:reasoning-process_diagram}
\end{figure*}

As shown in Fig.~\ref{fig:reasoning-process_diagram}, once MLLM receives input, it follows a structured reasoning process to select the next action. This process involves analysing candidate views, predicting future observations, analysing last action (More details list in Sec.~\ref {sec:Future-Past Bidirectional Reasoning}), evaluating stop conditions, and estimating safe moving parameters. 

The final output of each step is represented in a structured JSON format containing the reasoning trace, the selected view, the chosen action option, action parameters, and updated navigation history. This explicit format encourages the MLLM to maintain coherent reasoning and ensures interoperability of the decision-making process throughout the navigation episode.

\begin{itemize}
  \item \textbf{Thought}: Step-by-step reasoning trace of the current decision.
  \item \textbf{Selected Image}: The selected image index from the limited front views (1, 2, or 3).
  \item \textbf{Action Options}: A symbolic label (e.g., `A`) chosen from the available action candidates.
  \item \textbf{Degree}: Rotation degree if applicable (e.g., 30, 60), otherwise \texttt{null}.
  \item \textbf{Safe Distance}: A forward distance (in meters) the robot can safely travel in the selected direction, constrained by nearby obstacles.
  \item \textbf{Confuse}: A Boolean flag indicating whether the Disambiguation Module is triggered (\texttt{true} / \texttt{false}).
  \item \textbf{Updated History}: Structured record of navigation progress, including:
    \begin{itemize}
      \item \textbf{Trajectory Summary}: a textual summary of the current position and past movements.
      \item \textbf{Instruction Progress}: the updated subgoal completion status (\texttt{Completed} / \texttt{In Progress} / \texttt{Not Started}).
    \end{itemize}
\end{itemize}
To avoid getting stuck, the robot compares the current spatial descriptions $\mathcal{L}$ with the previous ones, and performs a slight rightward shift if they are unchanged.

\subsection{Uncertainty-Aware Reasoning}
\label{sec:uncertainty-aware}

In real-world navigation, the robot often encounters challenges like vague instructions, sensor noise, or conflicting visual and language inputs. These uncertainties can lead to poor decisions or cause the robot to get stuck in local areas. To enhance robustness and decision quality under such uncertainty, we propose an \textbf{Uncertainty-Aware Reasoning} module, which equips the robot with the ability to detect uncertainty and reason over both future and past context.

This module consists of two complementary mechanisms:
\begin{itemize}
    \item \textbf{Disambiguation Module}: explicitly detects ambiguous and triggers re-evaluation to avoid poor local decisions.
    \item \textbf{Future-Past Bidirectional Reasoning (FPBR)}: encourages the robot to reason forward (simulate outcomes) and backward (reflect on past decisions) to ensure globally coherent navigation.
\end{itemize}

\subsubsection{Disambiguation Module}
\label{sec:confuse_module}

Navigation instructions may be ambiguous, under-specified, or visually contradictory with the current scene. To handle such cases, we propose a {Disambiguation Module} that performs an explicit uncertainty check. After the initial step, the robot invokes the MLLM to determine whether the robot is “confused”. The binary decision \texttt{Confuse: true} is triggered if any of the following conditions are met: (1) the instruction is ambiguous, (2) the goal is unclear, or (3) the visual cues conflict with the expected progression.

When confusion is detected, the robot performs a full 360° rotation to collect panoramic observations, consisting of 12 RGB images $I^{\text{rgb}}_{1...12}$, semantic descriptions $\mathcal{O}^{(12)}$, and spatial descriptions $\mathcal{L}^{(12)}_{\text{spatial}}$, same as the initial observation. It also incorporates contextual inputs updated from the previous step: a \texttt{Trajectory Summary} and \texttt{Instruction Progress}, which respectively describe the robot’s navigation history and the completed portions of the instruction.

We construct a structured prompt to guide the MLLM through reasoning over this information. The prompt instructs the model to:
\begin{enumerate}
    \item Identify completed steps in the instruction.
    \item Detect any misalignment between the current heading and the instruction intent.
    \item Recommend a re-orientation direction and safe distance to resume navigation.
\end{enumerate}
The MLLM returns a structured JSON object:
\begin{itemize}
    \item \textbf{Selected Image}: The index of the RGB image that best represents the direction the robot should ini- tially face and begin moving toward.
    \item \textbf{Safe Distance}: A forward distance (in meters) the robot can safely travel in the selected direction, constrained by nearby obstacles.
\item \textbf{Updated History}: Structured record of navigation progress, including:
    \begin{itemize}
      \item \textbf{Trajectory Summary}: a textual summary of the current position and past movements.
      \item \textbf{Instruction Progress}: the updated subgoal completion status (\texttt{Completed} / \texttt{In Progress} / \texttt{Not Started}).
    \end{itemize}
\end{itemize}

\subsubsection{Future-Past Bidirectional Reasoning}
\label{sec:Future-Past Bidirectional Reasoning}

To maintain temporal consistency and improve long-horizon planning, we introduce a {Future-Past Bidirectional Reasoning (FPBR)} mechanism. Inspired by Local Observation-Action Reasoning~\cite{qiao2025navbench}, this component prompts the model to reason about possible future states while reflecting on past actions.

\noindent\textbf{Future Prediction:}
The robot is prompted to mentally simulate the visual consequences of alternative navigational actions (e.g., “If I turn left, I expect to see a hallway leading to a kitchen”). This simulation is conditioned on current RGB and semantic inputs, surrounding spatial descriptions, and any landmarks extracted from the instruction. The model evaluates the plausibility of each action by comparing expected visual outcomes with the goal context.

\noindent\textbf{Past Recall:}
The model also incorporates the \texttt{Previous Selected Image} and \texttt{Previous Thought} to recall its previous decision. It compares the current observation at step~$t$ with what it predicted at step~$t-1$. If a mismatch is detected (e.g., expected to see the “chicken” but sees a “bed”), the model may revise its policy accordingly.

This bidirectional reasoning promotes both short-term correction and long-term instruction alignment. By grounding each decision in both simulated futures and reflective pasts, the robot avoids cascading errors and achieves more coherent navigation trajectories.

\section{Experiments}

\subsection{Experiment Setup}

\begin{table*}[t]
\centering
\vspace{15pt}
\caption{Comparison of Real-World Navigation Efficiency and Performance on Hello Robot}
\label{tab:merged_latency_performance}
\renewcommand{\arraystretch}{1.2}
\resizebox{\linewidth}{!}{
\begin{tabular}{lcccccc}
\toprule
\textbf{Method} & \textbf{View Type} & \textbf{Perception Time (s)} & \textbf{Inference Time (s)} & \textbf{Total Time (s)} & \textbf{SR$\uparrow$} & \textbf{NE$\downarrow$} \\
\midrule
RecBERT~\cite{hong2022bridging}     & Panoramic     & 22.40  & 0.02  & 22.42 & 20          & 3.92 \\
SmartWay~\cite{smartway}            & Panoramic      & 22.40  & 6.85  & 29.25 & 32          & 3.01 \\
\midrule
\textbf{Ours}              & Frontal View   & 5.13   & 7.26  & 12.39 & \textbf{36}& \textbf{2.78}\\
\bottomrule
\end{tabular}}
\end{table*}

\begin{table*}[t]
\caption{Comparison on simulated Environment in R2R-CE dataset}
\label{tab:table_performance_comparison_r2r}
\vspace{-15pt}
\begin{center}
\resizebox{\linewidth}{!}{
\begin{tabular}{l|c|ccccc}
\toprule
\textbf{Method} & \textbf{View Type} & \textbf{TL} & \textbf{NE}$\downarrow$& \textbf{nDTW}$\uparrow$  &  \textbf{SR}$\uparrow$ & \textbf{SPL}$\uparrow$ \\
\midrule
\rowcolor{Grey!20}\multicolumn{7}{c}{\textbf{Supervised}}\\
CMA\cite{hong2022bridging}& Panoramic& 11.08 & 6.92&50.77  &  37 & 32.17 \\
RecBERT\cite{hong2022bridging}& Panoramic         & 11.06 & 5.80   &54.81&  48 & 43.22 \\
BEVBert\cite{an2022bevbert}& Panoramic               & 13.63 & 5.13 &61.40&  60 & 53.41 \\
ETPNav\cite{an2024etpnav}& Panoramic                 & 11.08 & 5.15 &61.15& 52 & 52.18 \\
\midrule
RecBERT\cite{hong2022bridging}& Frontal View & 7.89&  7.78&40.78&6 & 5.80\\
Navid~\cite{zhang2024navid}&Frontal View & 9.98& 7.24&-&28& 21.40\\

\midrule
\rowcolor{Cerulean!20}\multicolumn{7}{c}{\textbf{Zero-Shot}} \\
Random& Panoramic                 & 8.15  & 8.63&34.08    & 2  & 1.50 \\
LXMERT\cite{tan2019lxmert}& Panoramic & 15.79 & 10.48&18.73   & 2  & 1.87 \\
MapGPT-CE-GPT4o~\cite{chen2024mapgpt}& Panoramic &12.63&8.16&27.38&7&5.04\\
DiscussNav-GPT4\cite{long2023discuss}& Panoramic             & 6.27 & 7.77 &42.87   & 11 & 10.51 \\
Open-Nav-Llama3.1\cite{opennav}& Panoramic         & 8.07  & {7.25}  &44.99 & 16 & 12.90 \\
Open-Nav-GPT4\cite{opennav}& Panoramic         & 7.68  & {6.70}&45.79  & {19} & {16.10} \\

Smartway\cite{smartway}& Panoramic& 16.01 ± 0.85 & 6.81 ± 0.38&41.77 ± 2.42 & \textbf{29.00 ± 2.94} &22.08 ± 2.92 \\
\midrule
Random&Frontal View& 4.81 ± 0.19 &9.24 ± 0.07&33.22 ± 0.22  &1.25 ± 0.96& 1.18 ± 0.88 \\				
Smartway\cite{smartway}&Frontal View &10.91 ± 0.49&8.50 ± 0.28&41.19 ± 0.69&7.25 ± 3.77 & 6.32 ± 3.16\\
				
\textbf{Ours}&Frontal View & 12.56 ± 0.71&7.72 ± 0.42&\textbf{51.83 ± 1.54}&27.75 ± 2.22&\textbf{24.95 ± 2.70}\\

\bottomrule
\end{tabular}}
\vspace{-10pt}
\end{center}
\end{table*}

\noindent\textbf{Implementation Details} For zero-shot navigation, the MLLM-based navigator is deployed via the GPT-4o-2024-08-06 API, which is the same as baseline~\cite{smartway}. The occupancy distance thresholds are set as $d_{\text{close}} = 0.5$ and $d_{\text{mid}} = 4$. Due to the randomness in MLLM outputs, each experiment is run four times, and we report the average.

\noindent\textbf{Environments in the Simulator and Real World}
Our simulator implementation is based on the Habitat simulator~\cite{savva2019habitat}. Following the evaluation protocol of Open-Nav~\cite{opennav}, we assess our navigator with the same 100 episodes to ensure direct comparability with prior work.

For real-world validation, we deploy our system on the Hello Robot platform~\cite{HelloRobot2025} to evaluate per-step latency and deployment efficiency, while navigation performance is also measured exclusively on the robot. The Hello Robot is equipped with an Intel RealSense D435if RGB-D camera, and all onboard inference is executed on a workstation with the NVIDIA RTX 4090 GPU during operation.

To assess the system’s generalisation in real-world settings, following the experimental setup from SmartWay~\cite{smartway} and DiscussNav~\cite{long2023discuss}, we collect 25 diverse trajectories across multiple rooms, including open-vocabulary target landmarks (e.g., green bin). These trajectories vary in instruction complexity and spatial layout, enabling a comprehensive evaluation of our zero-shot navigation policy under realistic conditions.

\noindent\textbf{Evaluation Metrics}
We evaluate our navigator against several learning-based and zero-shot VLN methods using standard VLN metrics, including Success Rate (SR), normalized Dynamic Time Warping (nDTW), Success weighted by Path Length (SPL), Trajectory Length (TL), and Navigation Error (NE). Following prior works~\cite{opennav,smartway}, we consider a navigation to be successful if the robot stops within 3 meters of the goal in the simulator, and within 2 meters in real-world robot experiments.

\subsection{Results in Real-world Environments}

Table~\ref{tab:merged_latency_performance} compares our method with two existing baselines: a supervised approach (RecBERT~\cite{hong2022bridging}) and a zero-shot method (SmartWay~\cite{smartway}). Both baselines rely on panoramic inputs and operate on the Hello Robot platform~\cite{HelloRobot2025}.
RecBERT achieves the lowest inference time (0.02s), but its overall performance is limited, with a success rate (SR) of only 20 and a navigation error (NE) of 3.92. SmartWay improves these metrics (SR = 32, NE = 3.01), but its inference time increases to 6.85s, leading to the highest overall latency of 29.25s.
Here, the perception time refers to the time spent by the real-world robot on collecting visual observations, while the inference time is the average model prediction time per step.

In contrast, our method leverages only the frontal-view images, significantly reducing the perception time to 5.13s. Although its inference time is slightly higher (7.26s), the total latency is reduced to 12.39s, only 42.4\% of SmartWay’s total time. 
More critically, our method achieves the best navigation performance with the highest success rate (SR = 36) and the lowest navigation error (NE = 2.78), demonstrating its strong effectiveness under real-world constraints.

These results highlight that our approach, relying solely on front-view inputs, accelerates deployment and achieves strong navigation performance, demonstrating its practicality and scalability for real-world robotic navigation tasks. 

\subsection{Results in Simulator}

As shown in Table~\ref{tab:table_performance_comparison_r2r}, our method achieves strong zero-shot navigation performance on the R2R-CE benchmark, despite relying solely on frontal views rather than full panoramic inputs. Among panoramic-view baselines, Open-Nav-GPT4~\cite{opennav} and SmartWay~\cite{smartway}  perform well with SR of 19 and 29, and SPL of 16.10 and 22.08, respectively. In comparison, our method achieves a competitive SR of 27.75 and surpasses both baselines in SPL (24.95) and nDTW of 51.83 (vs. 41.77 for SmartWay and 45.79 for Open-Nav-GPT4). For fair comparison, we reproduce frontal-view baselines using our own implementation.

Overall, the results show that even with less visual input, our model performs competitively with panoramic-based methods.
These findings demonstrate that our frontal-view approach offers a strong balance of efficiency and effectiveness, making it well-suited for real-world deployment.

\subsection{Ablation Study}
\begin{table}[t]
\vspace{5pt}
\caption{Ablation study on the impact of the Disambiguation Module and Future-Past Bidirectional Reasoning (FPBR).}
\label{tab:ablation_study}
\begin{center}
\setlength\tabcolsep{2pt}
\resizebox{\linewidth}{!}{
\begin{tabular}{l|ccccc}
\toprule
\textbf{Method}  & \textbf{TL} & \textbf{NE}$\downarrow$ & \textbf{nDTW}$\uparrow$ & \textbf{SR}$\uparrow$ & \textbf{SPL}$\uparrow$ \\
\midrule
Ours (w/o Disambiguation and FPBR)  & 11.69 ± 0.63 & 7.64 ± 0.47	&50.83 ± 1.83&	19.75 ± 4.19&	17.42 ± 2.49 \\
+ Disambiguation only & 12.21 ± 0.45&	7.64 ± 0.49	&\textbf{52.06 ± 1.42}	&	24.25 ± 1.26&	21.14 ± 1.77\\
Full Model (with Disambiguation + FPBR)  & 12.56 ± 0.71&7.72 ± 0.42&51.83 ± 1.54&\textbf{27.75 ± 2.22}&\textbf{24.95 ± 2.70}\\
\bottomrule
\end{tabular}
}
\end{center}
\vspace{-15pt}
\end{table}

Table~\ref{tab:ablation_study} shows the ablation results on the R2R-CE dataset. When both the Disambiguation module and Future-Past Bidirectional Reasoning (FPBR) are removed, performance drops significantly (SR: 19.75, SPL: 17.42). Adding only the Disambiguation module boosts SR to 24.25 and SPL to 21.14. With both modules included, the full model achieves the best results (SR: 27.75, SPL: 24.95), confirming that each component contributes to stronger and more reliable navigation.

\section{Conclusion}
We presented Fast-SmartWay, a zero-shot VLN-CE framework that removes the need for panoramic view and waypoint predictors. Instead, it uses only three frontal views and a multimodal large language model to directly predict actions. To improve robustness and planning consistency, we introduced an Uncertainty-Aware Reasoning module that combines disambiguation and future-past bidirectional reasoning. Experiments in both simulation and real-world settings show that Fast-SmartWay achieves performance comparable to state-of-the-art (SOTA) panoramic view zero-shot methods, while offering much faster processing time. Our results demonstrate the potential of efficient end-to-end multimodal reasoning for scalable and practical navigation systems in real-world scenarios.


\bibliographystyle{IEEEtran}
\bibliography{clean_ref}

\begin{thebibliography}{10}
\providecommand{\url}[1]{#1}
\csname url@samestyle\endcsname
\providecommand{\newblock}{\relax}
\providecommand{\bibinfo}[2]{#2}
\providecommand{\BIBentrySTDinterwordspacing}{\spaceskip=0pt\relax}
\providecommand{\BIBentryALTinterwordstretchfactor}{4}
\providecommand{\BIBentryALTinterwordspacing}{\spaceskip=\fontdimen2\font plus
\BIBentryALTinterwordstretchfactor\fontdimen3\font minus \fontdimen4\font\relax}
\providecommand{\BIBforeignlanguage}[2]{{%
\expandafter\ifx\csname l@#1\endcsname\relax
\typeout{** WARNING: IEEEtran.bst: No hyphenation pattern has been}%
\typeout{** loaded for the language `#1'. Using the pattern for}%
\typeout{** the default language instead.}%
\else
\language=\csname l@#1\endcsname
\fi
#2}}
\providecommand{\BIBdecl}{\relax}
\BIBdecl

\bibitem{anderson2018vision}
P.~Anderson, Q.~Wu, D.~Teney, J.~Bruce, M.~Johnson, N.~S{\"u}nderhauf, I.~Reid, S.~Gould, and A.~Van Den~Hengel, ``Vision-and-language navigation: Interpreting visually-grounded navigation instructions in real environments,'' in \emph{CVPR}, 2018, pp. 3674--3683.

\bibitem{wang2023scaling}
Z.~Wang, J.~Li, Y.~Hong, Y.~Wang, Q.~Wu, M.~Bansal, S.~Gould, H.~Tan, and Y.~Qiao, ``Scaling data generation in vision-and-language navigation,'' in \emph{ICCV}, 2023, pp. 12\,009--12\,020.

\bibitem{hong2020recurrent}
Y.~Hong, Q.~Wu, Y.~Qi, C.~Rodriguez-Opazo, and S.~Gould, ``A recurrent vision-and-language bert for navigation,'' in \emph{CVPR}, June 2021, pp. 1643--1653.

\bibitem{chen2021hamt}
S.~Chen, P.-L. Guhur, C.~Schmid, and I.~Laptev, ``History aware multimodal transformer for vision-and-language navigation,'' \emph{NeurIPS}, vol.~34, pp. 5834--5847, 2021.

\bibitem{qiao2022hop}
Y.~Qiao, Y.~Qi, Y.~Hong, Z.~Yu, P.~Wang, and Q.~Wu, ``Hop: History-and-order aware pre-training for vision-and-language navigation,'' in \emph{CVPR}, 2022, pp. 15\,418--15\,427.

\bibitem{krantz2020beyond}
J.~Krantz, E.~Wijmans, A.~Majumdar, D.~Batra, and S.~Lee, ``Beyond the nav-graph: Vision-and-language navigation in continuous environments,'' in \emph{ECCV}.\hskip 1em plus 0.5em minus 0.4em\relax Springer, 2020, pp. 104--120.

\bibitem{hong2022bridging}
Y.~Hong, Z.~Wang, Q.~Wu, and S.~Gould, ``Bridging the gap between learning in discrete and continuous environments for vision-and-language navigation,'' in \emph{CVPR}, 2022, pp. 15\,439--15\,449.

\bibitem{an2022bevbert}
D.~An, Y.~Qi, Y.~Li, Y.~Huang, L.~Wang, T.~Tan, and J.~Shao, ``Bevbert: Topo-metric map pre-training for language-guided navigation,'' \emph{ICCV}, 2023.

\bibitem{an2024etpnav}
D.~An, H.~Wang, W.~Wang, Z.~Wang, Y.~Huang, K.~He, and L.~Wang, ``Etpnav: Evolving topological planning for vision-language navigation in continuous environments,'' \emph{TPAMI}, 2024.

\bibitem{li2025ground}
Z.~Li, G.~Zhou, H.~Hong, Y.~Shao, W.~Lyu, Y.~Qiao, and Q.~Wu, ``Ground-level viewpoint vision-and-language navigation in continuous environments,'' in \emph{ICRA}, 2025.

\bibitem{krantz2022sim}
J.~Krantz and S.~Lee, ``Sim-2-sim transfer for vision-and-language navigation in continuous environments,'' in \emph{ECCV}.\hskip 1em plus 0.5em minus 0.4em\relax Springer, 2022, pp. 588--603.

\bibitem{touvron2023llama}
H.~Touvron \emph{et~al.}, ``Llama: Open and efficient foundation language models,'' \emph{arXiv preprint arXiv:2302.13971}, 2023.

\bibitem{openai2023gpt4}
J.~Achiam, S.~Adler, S.~Agarwal, L.~Ahmad, I.~Akkaya, F.~L. Aleman, D.~Almeida, J.~Altenschmidt, S.~Altman, S.~Anadkat \emph{et~al.}, ``Gpt-4 technical report,'' \emph{arXiv preprint arXiv:2303.08774}, 2023.

\bibitem{zhou2024navgpt}
G.~Zhou, Y.~Hong, and Q.~Wu, ``Navgpt: Explicit reasoning in vision-and-language navigation with large language models,'' in \emph{AAAI}, vol.~38, no.~7, 2024, pp. 7641--7649.

\bibitem{opennav}
Y.~Qiao, W.~Lyu, H.~Wang, Z.~Wang, Z.~Li, Y.~Zhang, M.~Tan, and Q.~Wu, ``Open-nav: Exploring zero-shot vision-and-language navigation in continuous environment with open-source llms,'' in \emph{ICRA}, 2025.

\bibitem{smartway}
X.~Shi, Z.~Li, W.~Lyu, J.~Xia, F.~Dayoub, Y.~Qiao, and Q.~Wu, ``Smartway: Enhanced waypoint prediction and backtracking for zero-shot vision-and-language navigation,'' in \emph{IROS}, 2025.

\bibitem{qiao2025navbench}
Y.~Qiao, H.~Hong, W.~Lyu, D.~An, S.~Zhang, Y.~Xie, X.~Wang, and Q.~Wu, ``Navbench: Probing multimodal large language models for embodied navigation,'' \emph{arXiv preprint arXiv:2506.01031}, 2025.

\bibitem{HelloRobot2025}
``Hello robot : Open source mobile manipulator for ai \& robotics,'' \url{https://hello-robot.com/}, 2025.

\bibitem{zhang2024vision}
Y.~Zhang, Z.~Ma, J.~Li, Y.~Qiao, Z.~Wang, J.~Chai, Q.~Wu, M.~Bansal, and P.~Kordjamshidi, ``Vision-and-language navigation today and tomorrow: A survey in the era of foundation models,'' \emph{arXiv preprint arXiv:2407.07035}, 2024.

\bibitem{qi2020reverie}
Y.~Qi, Q.~Wu, P.~Anderson, X.~Wang, W.~Y. Wang, C.~Shen, and A.~v.~d. Hengel, ``Reverie: Remote embodied visual referring expression in real indoor environments,'' in \emph{CVPR}, 2020, pp. 9982--9991.

\bibitem{anderson2020rxr}
A.~Ku, P.~Anderson, R.~Patel, E.~Ie, and J.~Baldridge, ``Room-across-room: Multilingual vision-and-language navigation with dense spatiotemporal grounding,'' in \emph{EMNLP}, 2020, pp. 4392--4412.

\bibitem{thomason2020cvdn}
J.~Thomason, M.~Murray, M.~Cakmak, and L.~Zettlemoyer, ``Vision-and-dialog navigation,'' in \emph{CoRL}, 2020, pp. 394--406.

\bibitem{long2023discuss}
Y.~Long, X.~Li, W.~Cai, and H.~Dong, ``Discuss before moving: Visual language navigation via multi-expert discussions,'' \emph{ICRA}, 2024.

\bibitem{zhang2024navid}
J.~Zhang, K.~Wang, R.~Xu, G.~Zhou, Y.~Hong, X.~Fang, Q.~Wu, Z.~Zhang, and H.~Wang, ``Navid: Video-based vlm plans the next step for vision-and-language navigation,'' \emph{arXiv preprint arXiv:2402.15852}, 2024.

\bibitem{wang2024sim}
Z.~Wang, X.~Li, J.~Yang, Y.~Liu, and S.~Jiang, ``Sim-to-real transfer via 3d feature fields for vision-and-language navigation,'' \emph{arXiv preprint arXiv:2406.09798}, 2024.

\bibitem{Zhang2023RecognizeAA}
Y.~Zhang, X.~Huang, J.~Ma, Z.~Li, Z.~Luo, Y.~Xie, Y.~Qin, T.~Luo, Y.~Li, S.~Liu \emph{et~al.}, ``Recognize anything: A strong image tagging model,'' \emph{ArXiv}, vol. abs/2306.03514, 2023.

\bibitem{tan2019lxmert}
H.~Tan and M.~Bansal, ``Lxmert: Learning cross-modality encoder representations from transformers,'' in \emph{EMNLP-IJCNLP}, 2019, pp. 5100--5111.

\bibitem{chen2024mapgpt}
J.~Chen, B.~Lin, R.~Xu, Z.~Chai, X.~Liang, and K.-Y.~K. Wong, ``Mapgpt: Map-guided prompting for unified vision-and-language navigation,'' in \emph{ACL}, 2024.

\bibitem{savva2019habitat}
M.~Savva, A.~Kadian, O.~Maksymets, Y.~Zhao, E.~Wijmans, B.~Jain, J.~Straub, J.~Liu, V.~Koltun, J.~Malik \emph{et~al.}, ``Habitat: A platform for embodied ai research,'' in \emph{ICCV}, 2019, pp. 9339--9347.

\end{thebibliography}

\end{document}